\documentclass[conference]{IEEEtran}
\IEEEoverridecommandlockouts
\usepackage{cite}
\usepackage{amsmath,amssymb,amsfonts}
\usepackage{algorithmic}
\usepackage{graphicx}
\usepackage{textcomp}
\usepackage{xcolor}
\usepackage{subcaption}

\def\BibTeX{{\rm B\kern-.05em{\sc i\kern-.025em b}\kern-.08em
    T\kern-.1667em\lower.7ex\hbox{E}\kern-.125emX}}
\begin{document}

\title{Dynamic Dropout: Leveraging Conway's Game of Life for Neural Networks Regularization\\

\thanks{This work is partially funded by project PID2021-122402OB-C22/MICIU/AEI
/10.13039/501100011033 FEDER, UE and by the ACIISI-Gobierno de Canarias and European FEDER funds under project ULPGC Facilities Net and Grant \mbox{EIS 2021 04}.}
}

\author{\IEEEauthorblockN{David Freire-Obregón}
\IEEEauthorblockA{\textit{SIANI, ULPGC} \\
Spain \\
0000-0003-2378-4277}
\and
\IEEEauthorblockN{José Salas-Cáceres}
\IEEEauthorblockA{\textit{SIANI, ULPGC} \\
Spain \\
0009-0004-7543-3385}
\and
\IEEEauthorblockN{Modesto Castrillón-Santana}
\IEEEauthorblockA{\textit{SIANI, ULPGC} \\
Spain \\
0000-0002-8673-2725}
}

\maketitle

\begin{abstract}
Regularization techniques play a crucial role in preventing overfitting and improving the generalization performance of neural networks. Dropout, a widely used regularization technique, randomly deactivates units during training to introduce redundancy and prevent co-adaptation among neurons. Despite its effectiveness, dropout has limitations, such as its static nature and lack of interpretability. In this paper, we propose a novel approach to regularization by substituting dropout with Conway's Game of Life (GoL), a cellular automata with simple rules that govern the evolution of a grid of cells. We introduce dynamic unit deactivation during training by representing neural network units as cells in a GoL grid and applying the game's rules to deactivate units. This approach allows for the emergence of spatial patterns that adapt to the training data, potentially enhancing the network's ability to generalize. We demonstrate the effectiveness of our approach on the CIFAR-10 dataset, showing that dynamic unit deactivation using GoL achieves comparable performance to traditional dropout techniques while offering insights into the network's behavior through the visualization of evolving patterns. Furthermore, our discussion highlights the applicability of our proposal in deeper architectures, demonstrating how it enhances the performance of different dropout techniques.
\end{abstract}


\begin{figure}[htbp]
  \centering
  \begin{subfigure}[b]{0.35\textwidth}
    \centering
    \includegraphics[width=\textwidth]{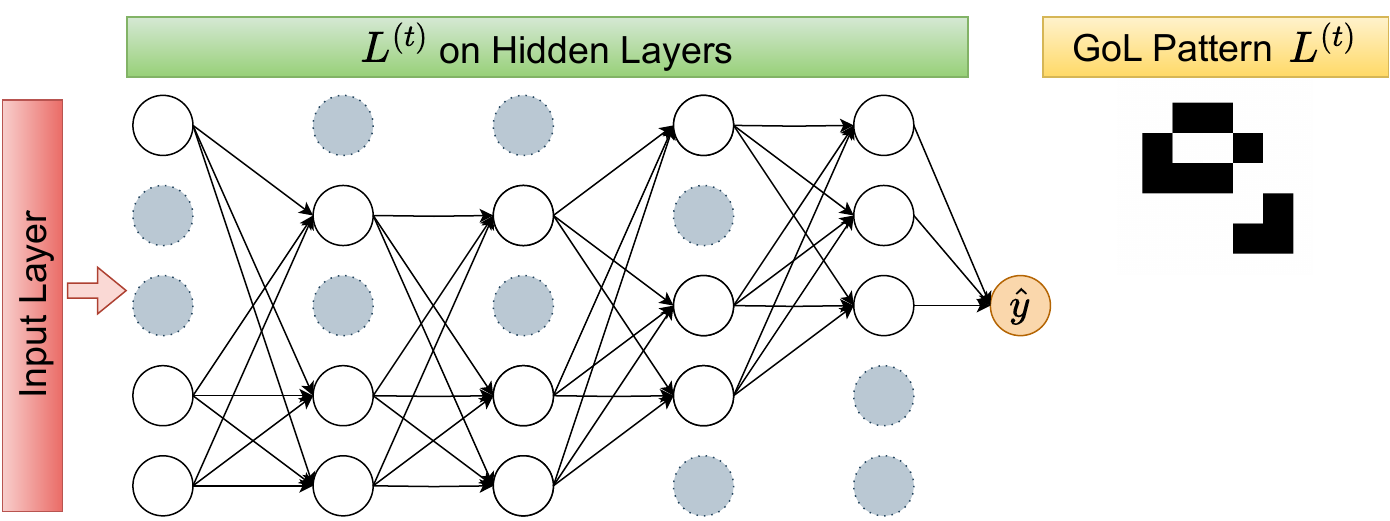}
    \caption{Dropout setting at epoch $t$}
  \end{subfigure}
  \hfill
  \begin{subfigure}[b]{0.35\textwidth}
    \centering
    \includegraphics[width=\textwidth]{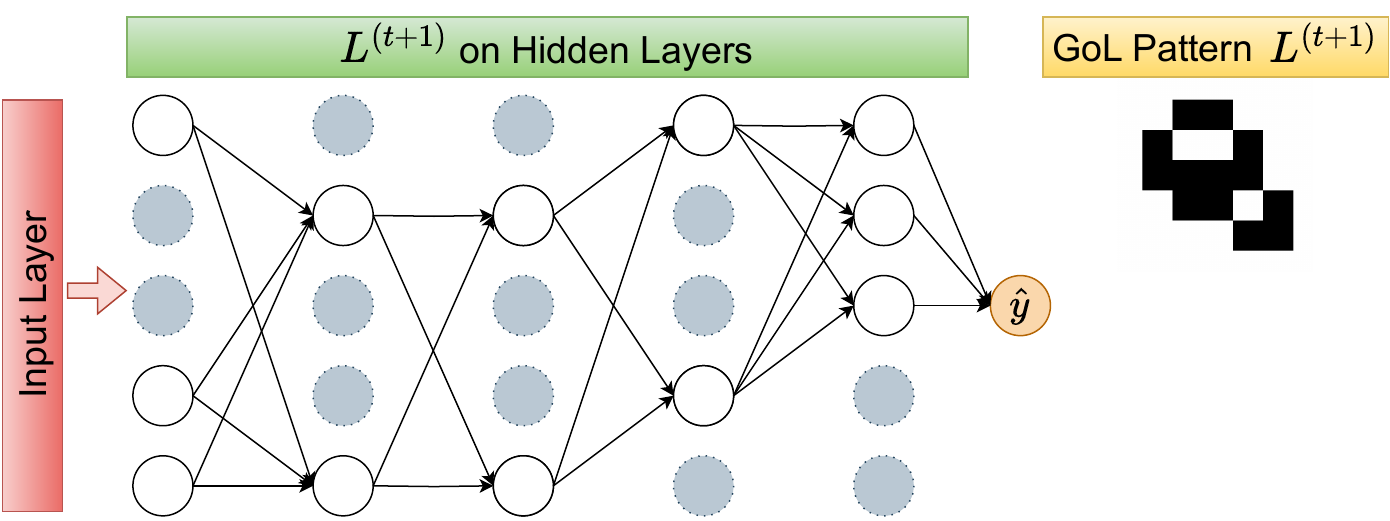}
    \caption{Dropout setting at epoch $t+1$}
  \end{subfigure}

  \vspace{0.8em} 

  \begin{subfigure}[b]{0.35\textwidth}
    \centering
    \includegraphics[width=\textwidth]{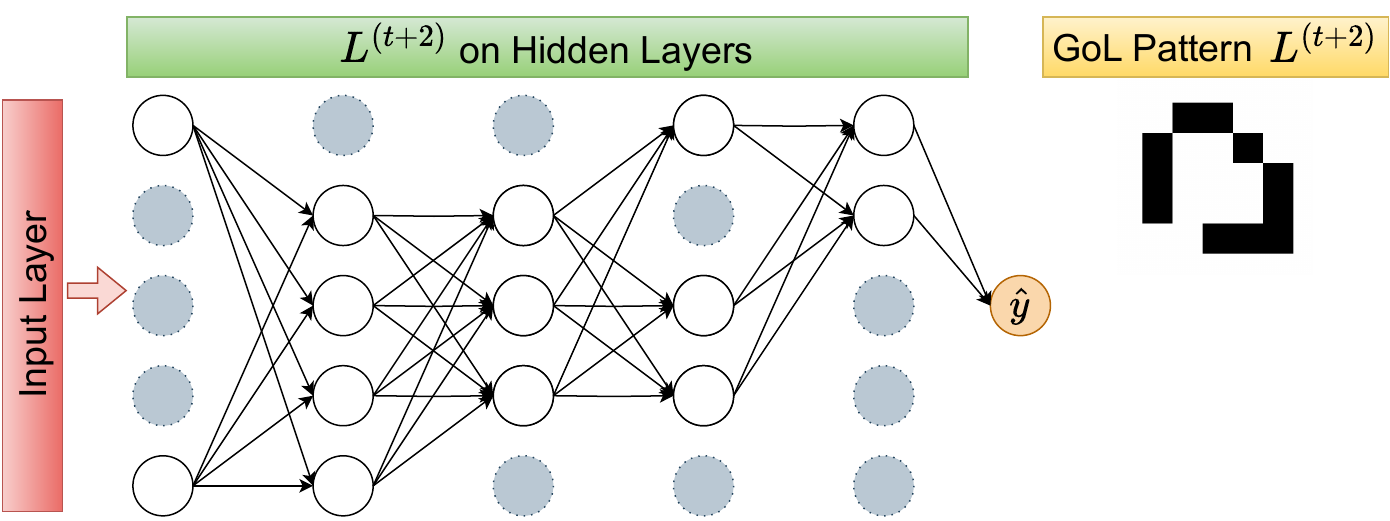}
    \caption{Dropout setting at epoch $t+2$}
  \end{subfigure}
  \hfill
  \begin{subfigure}[b]{0.35\textwidth}
    \centering
    \includegraphics[width=\textwidth]{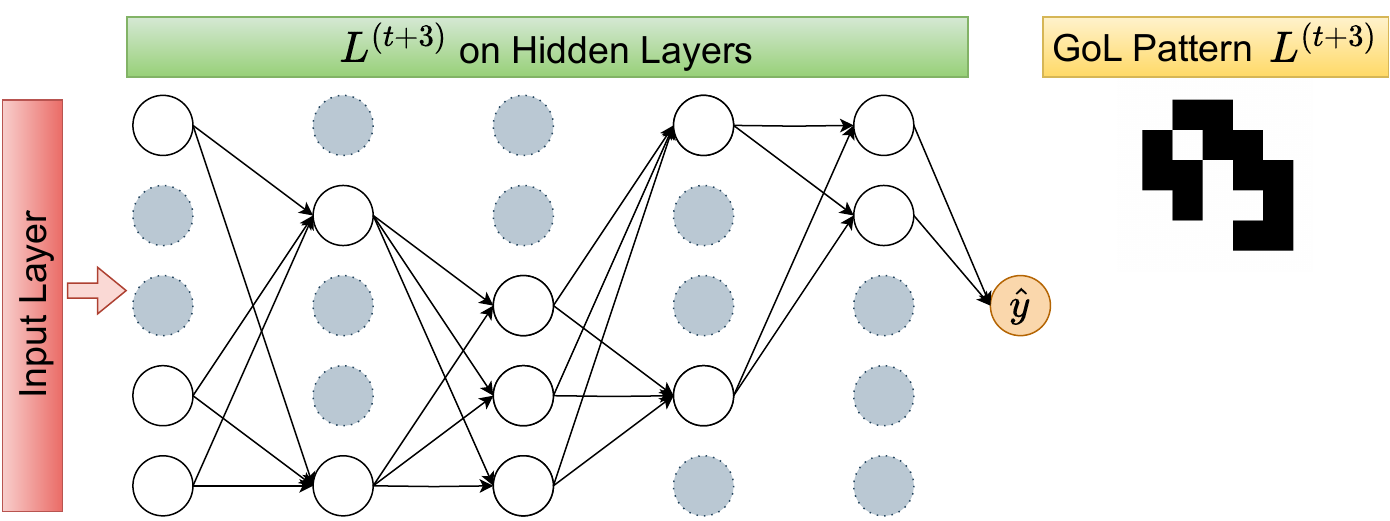}
    \caption{Dropout setting at epoch $t+3$}
  \end{subfigure}

  \caption{Dynamic Dropout at consecutive training epochs.}
  \label{fig:both}
\end{figure}

\section{Introduction}

Practical regression and classification face two opposite challenges: underfitting and overfitting \cite{Miyato19}. The latter is particularly critical, as models trained on limited data may minimize training error but still generalize poorly. Regularization addresses this issue by adding penalties that discourage overly complex solutions and improve robustness \cite{Yeom18}. 

Dropout is a widely used regularization method that randomly removes units and connections during training, effectively combining multiple network architectures \cite{Srivastava14}. Although effective, its stochastic and static nature limits adaptability to the network structure. 

In this work, we propose Dynamic Dropout, a self-organizing regularization method inspired by Conway’s Game of Life (GoL). Unlike random or Gaussian dropout, neuron activation depends on local neighborhood interactions, generating structured and adaptive sparsity patterns. This mechanism introduces spatial coherence and dynamic activation, offering a principled alternative that can improve generalization. 

Our main contributions are:
\begin{itemize}
    \item A GoL-driven dropout mechanism producing self-organizing activation patterns.
    \item A theoretical formulation including remarks on stability and computational efficiency.
    \item Empirical evaluation on CIFAR-10 with multiple dense architectures, demonstrating competitive generalization and interpretability.
\end{itemize}

The remainder of this paper reviews related regularization strategies, details our methodology, presents experiments, and concludes with key findings.

\section{Related Work}

The related work addresses two main themes: dropout regularization and the use of GoL principles in deep learning.

\textbf{Dropout}. Standard dropout prevents overfitting by randomly deactivating neurons during training \cite{Srivastava14}, encouraging redundant representations and improving robustness. Variants include \textit{DropConnect}, which drops connections instead of neurons \cite{Wan13}; \textit{Gaussian Dropout}, which adds Gaussian noise for improved generalization \cite{Wang13Fast}; and \textit{Alpha Dropout}, which preserves input statistics to stabilize training \cite{Klambauer17}. Dropout methods have proven effective in domains such as biometrics \cite{freire23prl}, segmentation \cite{Changlu19}, domain adaptation \cite{freire24mtap}, and expression recognition \cite{ojsantana22mtool}. Recent studies have proposed adaptive regularization strategies that dynamically adjust neuron behavior to the learning context. Similarly, agent-based systems have demonstrated self-organizing mechanisms that modulate activity and perception based on environmental feedback, reflecting principles consistent with our self-regulating dropout dynamics \cite{freire25fading, freire25wrong}.

\textbf{GoL}. Conway’s Game of Life (GoL) is a cellular automaton where cell states evolve based on neighbors, yielding complex emergent patterns \cite{Conway70}. Beyond theoretical interest, GoL has been applied to generative modeling \cite{Mordvintsev20}, cryptography \cite{Nandi94}, and vision tasks \cite{Yao15}. Recent work shows neural architectures can learn such dynamical rules directly \cite{Gilpin18}. Building on this, our study explores dropout guided by GoL-inspired neighbor interactions to regulate neural activation during training.

\section{Methodology}
\label{sec:metho}


\textbf{Model Architecture}.  The neural network comprises a Sequential model composed of an input layer, multiple hidden layers utilizing ReLU activation, and an output layer employing softmax activation.

Each dense layer computes:
\[
z^{(l)} = W^{(l)} a^{(l-1)} + b^{(l)}
\]
\[
a^{(l)} = \text{ReLU}(z^{(l)}) = \max(0, z^{(l)})
\]
Where \( W^{(l)} \) and \( b^{(l)} \) are the weights and biases of layer \( l \), and \( a^{(l-1)} \) is the activation from the previous layer.

The final layer's softmax function converts logits to $\eta$ class probabilities:
\[
p_c = \frac{e^{z_c}}{\sum_{k=1}^{\eta} e^{z_k}}
\]
where \( p_c \) is the probability of class \( c \), and \( z_c \) are the logits computed by the final layer.

\textbf{Dynamic Dropout}. The proposed regularization method is utilized, which depends on an evolving lattice. The lattice \( L \), representing the dropout mask, is a binary matrix with dimensions \( m \times q \), where \( m \) denotes the number of layers and \( q \) represents the units per layer in the neural network (see Figure \ref{fig:both}). The state of each cell \( L_{ij}^{(t+1)} \) at the next epoch is determined by the following rules, based on the number of active neighbors \( N_{ij} \):

\[
L_{ij}^{(t+1)} = 
\begin{cases} 
1 & \text{if } L_{ij}^{(t)} = 1 \text{ and } (N_{ij} = 2 \text{ or } N_{ij} = 3) \\
1 & \text{if } L_{ij}^{(t)} = 0 \text{ and } N_{ij} = 3 \\
0 & \text{otherwise}
\end{cases}
\]

Where:
\[
N_{ij} = \sum_{q=-1}^1 \sum_{r=-1}^1 L_{i+q, j+r}^{(t)} - L_{ij}^{(t)}
\]
This sum calculates the total number of active neighboring cells for \( L_{ij} \), excluding the cell itself.

Finally, each dense layer in the network performs a linear transformation followed by a non-linear activation function (ReLU), modified by a dynamic dropout mechanism determined by the lattice \( L \). The computation for each layer \( l \) can be described as follows:

\[
z^{(l)} = W^{(l)} a^{(l-1)} + b^{(l)}
\]
\[
\tilde{z}^{(l)} = z^{(l)} \odot (1 - L^{(t)}_{l})
\]
\[
a^{(l)} = \text{ReLU}(\tilde{z}^{(l)}) = \max(0, \tilde{z}^{(l)})
\]

Where:
\begin{itemize}
    \item \( W^{(l)} \) and \( b^{(l)} \) are the weight matrix and bias vector of the dense layer \( l \),
    \item \( a^{(l-1)} \) is the activation output from the previous layer \( l-1 \),
    \item \( L^{(t)}_{l} \) represents the dropout mask for layer \( l \) at a given epoch \( t \) , derived from the dynamic lattice. 
    Essentially, it signifies column \( l \) of the lattice at a specific epoch, with each element in the column corresponding to a unit in \( z^{(l)} \). Consequently, \( L^{(t)}_{l} \) has the same dimensions as the output of \( z^{(l)} \),
    \item \( \odot \) denotes element-wise multiplication,
    \item \( \tilde{z}^{(l)} \) represents the element-wise product of the linear output \( z^{(l)} \) and the dropout mask \( L^{(t)}_{l} \) , effectively deactivating certain neurons according to the lattice.
\end{itemize}

\begin{figure}[ht]
    \centering
    \includegraphics[scale=0.7]{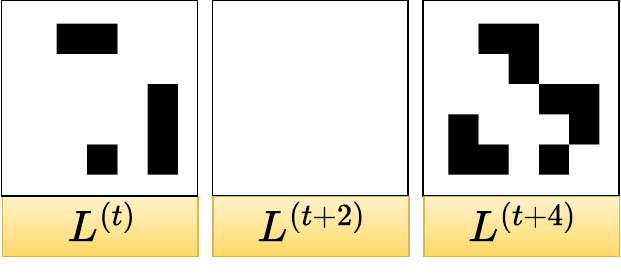}
    \caption{
        The evolution of GoL can sometimes result in fully activated configurations, which may contribute to overfitting, as observed in epoch $t+2$. To mitigate this, the algorithm randomly activates a subset of units, as depicted in epoch $t+4$, prompting a new iteration of GoL from that point onward.
    }
    \label{fig:overfitting}
\end{figure}

When overfitting is detected, a specific number of inactive units in the dropout mask \( L \) are randomly set to active (1); see Figure \ref{fig:overfitting}. This can be described by the random selection of indices where \( L_{ij} = 0 \) (inactive units) and setting them to 1 up to a predefined limit. Let \( P \) denote the number of units to activate, the update can be formalized as:
\[
L_{ij}^{(t+1)} = 
\begin{cases} 
1 & \text{if } (i, j) \in S \\
L_{ij}^{(t)} & \text{otherwise}
\end{cases}
\]
where \( S \) is a set of selected indices corresponding to \( P \) randomly chosen inactive units from \( L^{(t)} \). Integrating the Dynamic Dropout directly into the dense layer computation impacts the linear transformation and ReLU activation. The dropout mask \( L^{(l)} \) selectively deactivates neurons based on the state of the lattice, influencing the learning process by dynamically adjusting the network's complexity.

\textbf{Loss}. The categorical cross-entropy loss function measures the discrepancy between the true labels and the predicted probabilities. Assuming the output layer applies a softmax function to the activations from the last hidden layer, the loss for a single sample can be described as:

\[
L(y, \hat{y}) = -\sum_{c=1}^{C} y_c \log(\hat{y}_c)
\]

where \( \hat{y} \) are the predicted probabilities computed as:

\[
\hat{y} = \text{softmax}(z^{(\text{output})}) = \left[ \frac{e^{z_c^{(\text{output})}}}{\sum_{k=1}^{C} e^{z_k^{(\text{output})}}} \right]_{c=1}^{C}
\]

\[
z^{(\text{output})} = W^{(\text{output})} a^{(\text{last hidden})} + b^{(\text{output})}
\]

Here:
\begin{itemize}
    \item \( W^{(\text{output})} \) and \( b^{(\text{output})} \) are the weights and biases of the output layer,
    \item \( a^{(\text{last hidden})} \) represents the activations from the last hidden layer, which may have been modified by the Dynamic Dropout in earlier layers but not in the output layer itself.
\end{itemize}

$y_c$ is the true label for class \( c \) in one-hot encoding, and \( C \) is the number of classes.

This formulation reflects that the Dynamic Dropout does not affect the output layer but may influence the input to this layer through modifications in earlier layers. Thus, while Dynamic Dropout alters the intermediate activations that feed into the output layer, the final predictions and the associated loss are calculated without any direct Dynamic Dropout manipulation in the output layer itself.

One significant advantage of the proposed neural network architecture is its dynamic approach to managing dropout. Unlike classical techniques that randomly deactivate a fixed proportion of neurons, this method adapts activations according to the evolving GoL patterns. Neuron states are influenced by their local neighborhood, producing structured sparsity that reflects the network’s internal dynamics. As training progresses, lattice patterns typically stabilize after several epochs, ensuring consistent gradient flow without divergence. This evolving and context-dependent dropout pattern leads to more robust learning, as the network becomes less reliant on specific neurons and generalizes more effectively across features. The additional computational overhead is linear in the number of neurons and fully parallelizable. When overfitting is detected—identified by stagnation in validation loss—a small subset of inactive units is randomly reactivated to prevent lattice saturation and restore diversity in neuron participation.

\section{Experimental Setup}

\textbf{Dataset}. CIFAR-10 is a standard benchmark in computer vision, containing 60,000 $32 \times 32$ color images across ten classes, with 6,000 per class \cite{Krizhevsky09}. Its small size makes training efficient, while class diversity (including animals, vehicles, and household items) adds complexity. Images also vary in lighting, background, and viewpoint, enhancing realism. Due to its popularity and difficulty, CIFAR-10 is well-suited for evaluating regularization methods, such as dropout, which enables the assessment of generalization and robustness across diverse objects and scenes.

\begin{figure}[!ht]
    \centering
    \begin{subfigure}{0.3\textwidth}
        \centering
        \includegraphics[width=\linewidth]{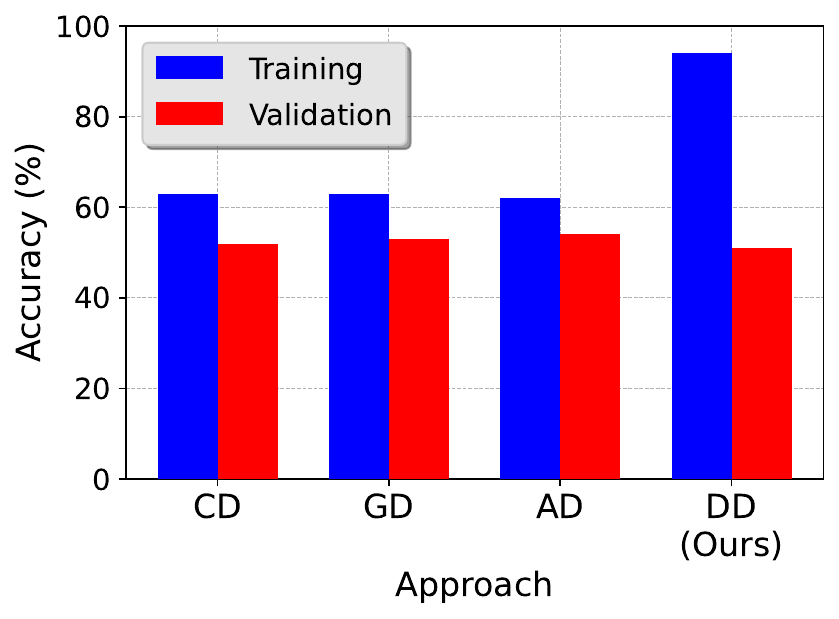}
        \caption{Architecture\_1 Accuracy}
    \end{subfigure}

    \begin{subfigure}{0.3\textwidth}
        \centering
        \includegraphics[width=\linewidth]{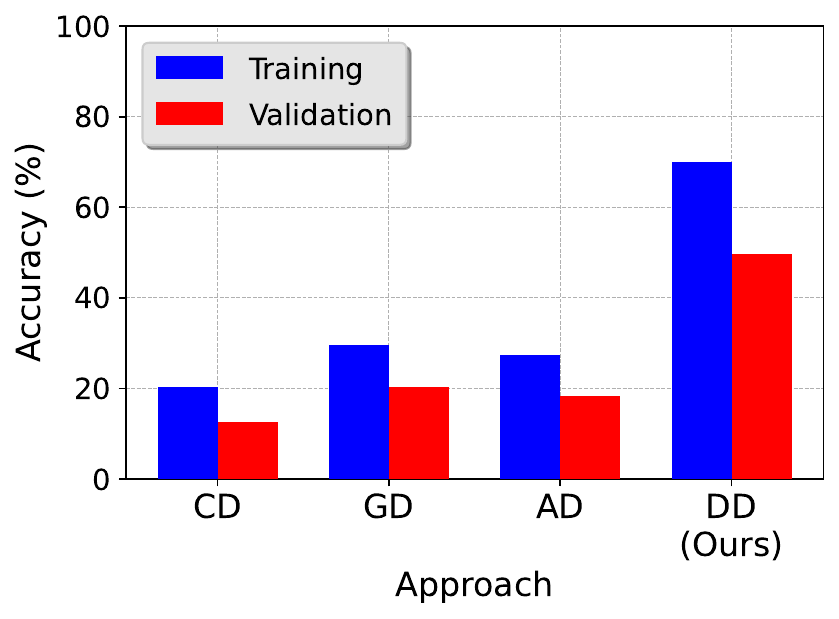}
        \caption{Architecture\_2 Accuracy}
    \end{subfigure}

    \begin{subfigure}{0.3\textwidth}
        \centering
        \includegraphics[width=\linewidth]{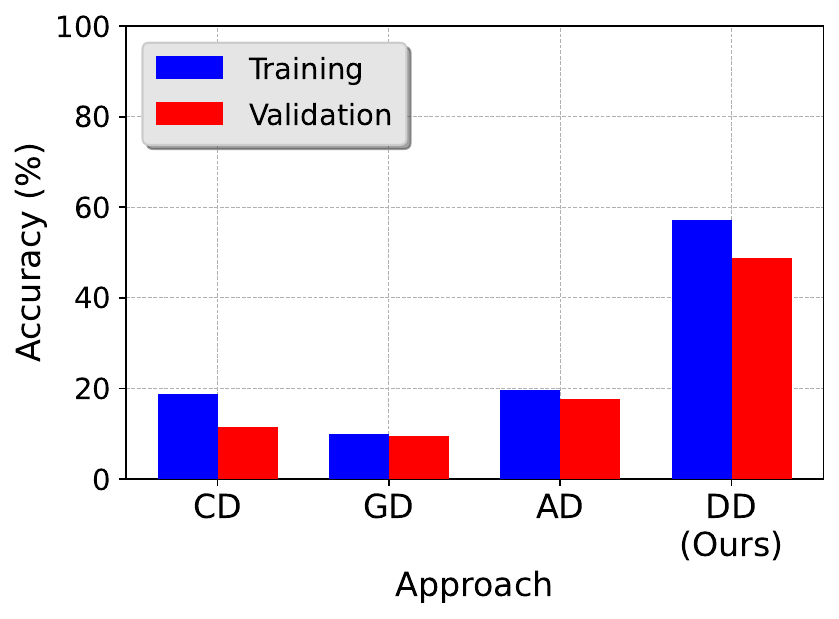}
        \caption{Architecture\_3 Accuracy}
    \end{subfigure}

    \caption{Accuracy achieved by each approach.}
    \label{fig:acc}
\end{figure}

\textbf{Metrics}. Evaluating Dynamic Dropout requires multiple metrics. Train and validation accuracy indicate the model’s learning and generalization capacity, while discrepancies reveal overfitting and the need for regularization. Train and test loss further quantify prediction errors, with low values reflecting robust performance. Finally, the generalization gap—differences between train and validation results highlights overfitting when significant, reinforcing the role of regularization.

\begin{figure}[!ht]
    \centering
    \begin{subfigure}{0.3\textwidth}
        \centering
        \includegraphics[width=\linewidth]{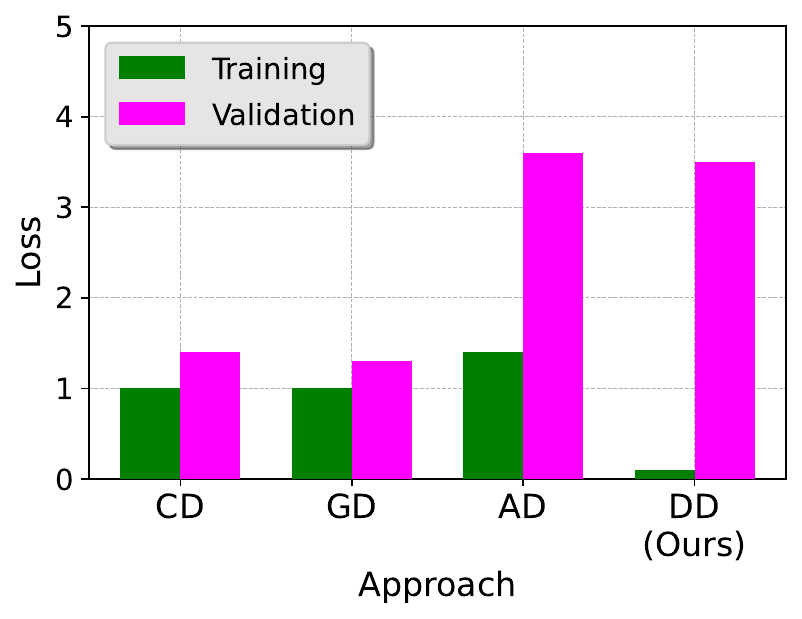}
        \caption{Architecture\_1 Loss}
    \end{subfigure}
    \hfill
    \begin{subfigure}{0.3\textwidth}
        \centering
        \includegraphics[width=\linewidth]{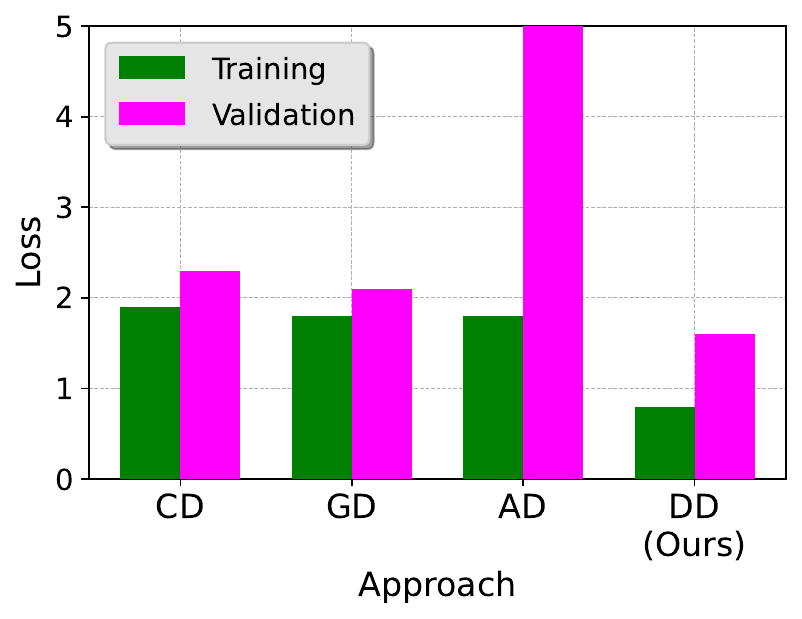}
        \caption{Architecture\_2 Loss}
    \end{subfigure}
    \hfill
    \begin{subfigure}{0.3\textwidth}
        \centering
        \includegraphics[width=\linewidth]{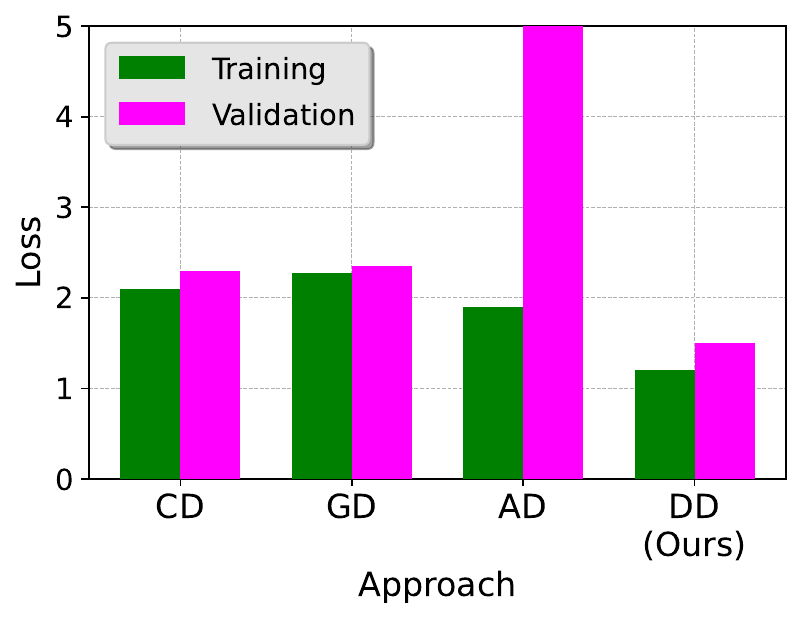}
        \caption{Architecture\_3 Loss}
    \end{subfigure}

    \caption{Loss achieved by each approach. AD validation losses in (b) and (c) exceeded the chart's display limits.}
    \label{fig:loss}
\end{figure}

\section{Experimental Evaluation}
We evaluated our proposed Dynamic Dropout (DD) mechanism against Classical Dropout (CD), Gaussian Dropout (GD), and Alpha Dropout (AD), following the configuration in \cite{Srivastava14} (input dropout rate of 0.5). Experiments were carried out on three architectures: Architecture\_1 (3 dense layers with 512 units each, representing a wide, shallow network), Architecture\_2 (10 dense layers with 128 units each), and Architecture\_3 (10 dense layers with 64 units each), trained for 100 epochs with a batch size of 512.

Across all configurations, DD achieved markedly higher training accuracies than the traditional methods. For example, in Architecture\_1 the training accuracy was approximately 62--63\% for CD, GD, and AD, while DD reached 94\%. However, DD's superior training performance did not fully extend to validation: in Architecture\_1, its validation accuracy was only 51\% with a loss of 3.5, suggesting some overfitting. Similar trends were seen in Architecture\_3, where DD attained 57.3\% training versus 48.9\% validation accuracy, and in deeper architectures (Architectures 2 and 3), where CD, GD, and AD struggled to exceed 30\% training accuracy, as shown in Figure~\ref{fig:acc}.

The generalization gap (the difference between training and validation accuracies) further illustrates these effects. In Architecture\_1, DD's 43\% gap (94\% vs.\ 51\%) contrasts with the more modest gaps of around 10--11\% observed for the other methods, while in Architecture\_2 DD showed a gap of 20.3\% (69.9\% training vs.\ 49.6\% validation). Although DD leverages architectural features to boost training performance significantly, its validation results indicate a tendency toward overfitting compared to traditional dropout techniques.

Architecture\_3 again demonstrates DD's ability to effectively minimize the generalization gap compared to other dropout methods. DD achieves a training accuracy of 57.3\% and a validation accuracy of 48.9\%, resulting in an 8.4\% generalization gap. This is markedly lower than the gaps exhibited by other methods (CD, GD, AD) which range up to around 11.2\%. This consistent reduction in the generalization gap across architectures highlights DD's robustness and effectiveness in diverse network settings, suggesting it not only learns more efficiently but also transfers this learning more effectively to unseen data.

Interestingly, as the architecture goes deeper, DD displays a marked reduction in overfitting tendencies. In these configurations, while still maintaining superior training performance, the gap between DD's training and validation accuracies decreases, and the validation losses become more competitive when compared to other methods. For example, in Architecture\_3, DD achieves a training accuracy of 57.3\% versus a validation accuracy of 48.9\%, which, while still indicative of some overfitting, shows a smaller disparity than in more narrowly architecture setups. Similarly, validation losses in these architectures for DD, though still on the higher side, are closer to those observed with other dropout methods, suggesting a better balance between learning and generalization as the network architecture becomes more complex, as detailed in Figure \ref{fig:loss}.

\begin{figure}[!ht]
    \centering
    \begin{subfigure}{0.25\textwidth}
        \centering
        \includegraphics[width=0.5\linewidth]{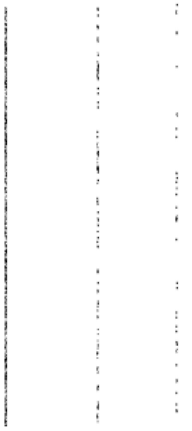}
        \caption{$L^{(1)}$, $L^{(10)}$, and $L^{(20)}$ in Architecture\_1}
    \end{subfigure}
    \hfill
    \begin{subfigure}{0.25\textwidth}
        \centering
        \includegraphics[width=0.5\linewidth]{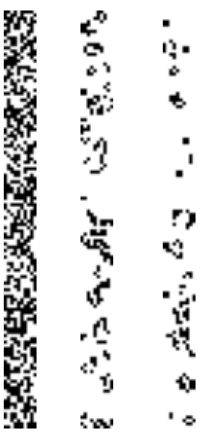}
        \caption{$L^{(1)}$, $L^{(10)}$, and $L^{(20)}$ in Architecture\_2}
    \end{subfigure}
    \hfill
    \begin{subfigure}{0.25\textwidth}
        \centering
        \includegraphics[width=0.5\linewidth]{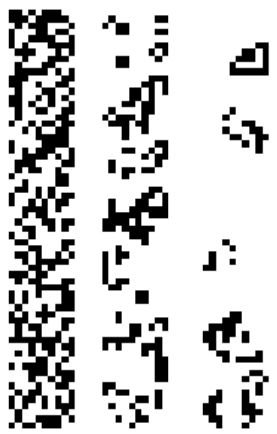}
        \caption{$L^{(1)}$, $L^{(10)}$, and $L^{(20)}$ in Architecture\_3}
    \end{subfigure}

    \caption{Layouts of the GoL lattice at epochs 1, 10, and 20 for the studied configurations.}
    \label{fig:configs}
\end{figure}

The observed behavior in DD can be attributed to the inherent characteristics of the GoL algorithm, which tends to perform more optimally in deeper and more square-like lattice configurations, see Figure \ref{fig:configs}. As the network's architecture widens (featuring more layers and units) DD's ability to modulate neuron activation becomes increasingly effective. This is likely because a broader lattice allows for a more nuanced application of GoL rules, facilitating a richer pattern of activations and deactivations that mirror complex feature interactions more accurately. This enhanced adaptability of DD in deeper networks results in a notable reduction in overfitting. The larger, more square-ish lattice structure provides a robust framework for dynamically adjusting neuron participation, thus aligning the dropout process more closely with the actual data distribution and feature relevance.
\section{Conclusions}
\label{sec:con}

This paper introduced Dynamic Dropout, a self-organizing regularization technique where neuron activation evolves according to Conway’s Game of Life. Replacing random deactivation with structured, context-dependent sparsity, the method improved training accuracy by up to 30\% over classical dropout and reduced overfitting in deeper networks. The approach adds negligible computational cost and remains fully parallelizable. Future work will extend this mechanism to CNNs and transformer architectures, incorporate adaptive thresholds, and integrate with batch normalization or Bayesian dropout for enhanced stability.

\bibliographystyle{IEEEtran}

\end{document}